\documentclass{llncs}

\usepackage{amsmath,amssymb}
\usepackage{booktabs}
\usepackage{color}
\usepackage{xspace}
\usepackage{tabularx}
\usepackage{graphicx}
\usepackage{siunitx}

\newcommand*{\eg}{e.g.\@\xspace}
\newcommand*{\ie}{i.e.\@\xspace}
\newcommand*{\cf}{cf.\@\xspace}
\newcommand*{\etal}{\textit{et al.}\@\xspace}

\makeatletter
\newcommand*{\etc}{%
    \@ifnextchar{.}%
        {etc}%
        {etc.\@\xspace}%
}
\makeatother

\begin{document}

\title{Assessment of Breast Cancer Histology using Densely Connected Convolutional Networks}
\titlerunning{\dots}  
%
\author{Matthias Kohl\inst{1} \and Christoph Walz\inst{2} \and Florian Ludwig\inst{1} \and Stefan Braunewell\inst{1} \and Maximilian Baust\inst{1}}
\authorrunning{Kohl et al.} 
%
%
\institute{Konica Minolta Laboratory Europe,\\
\email{maximilian.baust@konicaminolta.eu}, \texttt{http://research.konicaminolta.eu/}
\and
Institute of Pathology, Faculty of Medicine, LMU Munich}

\maketitle              

\begin{abstract}
Breast cancer is the most frequently diagnosed cancer and leading cause of cancer-related death among females worldwide.
In this article, we investigate the applicability of densely connected convolutional neural networks to the problems of histology image classification and whole slide image segmentation in the area of computer-aided diagnoses for breast cancer.
To this end, we study various approaches for transfer learning and apply them to the data set from the 2018 grand challenge on breast cancer histology images (BACH). 
\end{abstract}
\section{Introduction}
This work presents approaches for the classification of microscopy images as well as the segmentation of whole slide images (WSIs) in the area of computer-aided diagnosis for breast cancer.
In particular, it describes how the recently invented densely connected convolutional neural networks \cite{huang17} can be applied to the aforementioned tasks on data from the 2018 grand challenge in Breast Cancer Histology Images (BACH).
\subsubsection{Clinical Background}
According to the global cancer statistics 2012 \cite{Torre15}, breast cancer is the most frequently diagnosed cancer and the leading cause of cancer-related death among females worldwide, with an estimated 1.6 million new cases and over 0.5 million deaths per year.
With tumor stage remaining the most important determinant of the outcome \cite{warner11}, an early detection of breast cancer is crucial for reducing mortality rates.
Among other factors, such as patient  age,  axillary  lymph  node  status,  tumor  size, hormone receptor status, and HER2 status, histological features play an important role for categorizing patients with invasive breast cancer in order to assess prognosis and determine the appropriate therapy \cite{schnitt10}.
While the importance of histomorphological grading for breast cancer has been acknowledged almost 30 years ago \cite{elston91}, the computerized assessment of histological features has become increasingly popular during the last decade.
Tumor grading in breast cancer is typically based on the following three criteria suggested by Elston and Ellis \cite{elston91}:
\begin{enumerate}
	\item \textit{mitotic activity} as a measure of cellular proliferation,
	\item \textit{nuclear pleomorphism}, \ie how different the tumor cells are in comparison to normal cells, and
	\item \textit{glandular and tubular differentiation}, \ie how well the tumor resembles normal structures.
\end{enumerate}

Current developments in the area of digital pathology are driven by the observation that genetic and phenotypic intra-tumor heterogeneity have a direct impact on both diagnosis and disease management \cite{martelotto14} as well as the availability of effective machine learning techniques, such as deep convolutional neural networks.
Particularly the segmentation of WSIs, \ie the second part of the BACH challenge, plays an increasingly important role as it facilitates not only a standardized assessment of resection margins, but also novel scoring approaches, such as the ImmunoScore \cite{fridman12}, and a better understanding of tumor heterogeneity and micro-environment, \eg via phenotype-guided genetic readouts.
\subsubsection{Related Work}
in digital pathology can be categorized with respect to approaches which focus on the three aforementioned criteria for breast cancer grading as well as approaches for WSI segmentation.
In the following discussion we focus on the most recent approaches for beast cancer and breast cancer metastases that are based on deep learning.
For a more exhaustive overview, we refer the interested reader to recent overview articles, such as \cite{robertson17} or \cite{janowczyk16b}.

Possibly the largest class of methods focuses on the computational assessment of mitotic activity.
This field has been extensively promoted by the recent success of deep-learning-based approaches starting with the seminal work of \cite{ciresan13}. 
Referring the interested reader to the review paper of \cite{veta15} for an overview of all methods for mitosis detection until 2015, we specifically want to mention the more recent works on leveraging the potential of crowdsourcing for training deep networks \cite{albarqouni16}, on deep regression networks \cite{chen16} and on using deep residual Hough voting \cite{wollmann17b}.
The next category, comprises methods aiming at cellular or nuclear features.
Recent examples include works on stacked sparse auto encoders for nuclei detection \cite{xu16} and on hierarchical learning \cite{janowczyk16a}.
Regarding the assessment of glandular and tubular structures, there are only a few works in the field of breast cancer, such as \cite{dong14} or \cite{apou16}.
However, for a general overview on the computational assessment of relevant pathological structures and primitives, we refer the interested reader to \cite{janowczyk16b}.

In contrast to the approaches for particular histopathological tasks, there is the group of methods that are aiming at classification of whole tissue regions or at WSI-segmentation, which requires learning of features on both cellular and structural level.
A good example is the recent work of the BACH challenge organizers presenting a classification method for Hematoxylin- and Eosin-stained (HE-stained) histological images from breast cancer patients, \cf Araujo \etal \cite{araujo17}.

Regarding WSI-segmentation, there exists a series of methods is related to the recent challenges on cancer metastasis detection in lymph node (CAMELYON16 \& CAMELYON17).
Examples for notable works using the associated data sets are the ones of the organizers \cite{litjens16,bejnordi17}, as well as the works of Wang \etal \cite{wang16} or Liu \etal \cite{liu17}.
Conceptually, these approaches are also comparable to the recent works of Su \etal \cite{su15} and Cruz-Roa \etal \cite{cruz17}.
\subsubsection{Contributions and Organization}
We participated in the BACH challenge due to our interest in the learning and integration of features from multiple levels and their application to WSI-segmentation, particularly in case of small data sets.
As several contributions for the CAMELYON challenges were based on the popular Inception-v3 architecture proposed by \cite{szegedy16}, we wanted to assess the performance of another recently published and very promising architecture,\ie the densely connected convolutional networks proposed by Huang \etal \cite{huang17}.
As the data set of the challenge is too small for training such large architectures from scratch, we investigated two approaches for transfer learning: One based on weights obtained from training on ImageNet \cite{deng09} and one based one weights obtained from training the network on data from the CAMELYON challenges, which is described in Sec.~\ref{sec:methodology}.
The evaluation of these two approaches for both sub-challenges is described in Sec.~\ref{sec:evaluation} and discussed in Sec.~\ref{sec:discussion}, before we conclude this paper with Sec.~\ref{sec:conclusion}.

 \section{Methodology}
\label{sec:methodology}
The BACH challenge comprises two sub-challenges, \ie classification of histology images (part A) and segmentation of WSIs (part B).
In order to achieve these goals, we train classifiers $\mathcal{C}:p\mapsto\ell$ to predict the correct label $\ell$ for a given microscopic image (in case of part A) or a patch extracted from a WSI (in case of part B) $p$.
Thereby $\ell\in\{0,1,2,3\}$, where the numeric values encodes one of the four class labels: normal (0), benign (1), carcinoma \textit{in situ} (2),  invasive carcinoma (3).
As explained later in Sec.~\ref{sec:partA} and Sec.~\ref{sec:partB}, these classifiers are implemented via densely connected convolutional neural networks (DenseNets).
\subsection{Pre-training on CAMELYON data}
\label{sec:camelyon}
As the size of both data sets for part A and part B is too small for training deep networks from scratch, we decided to employ transfer learning with pre-trained networks.
Besides using a network which has been pre-trained on ImageNet data \cite{deng09}, we also investigated the possibility of using a network pre-trained on data from the two CAMELYON challenges \cite{bejnordi17}.
As of now, the data from these challenges consists of approximately 691 WSIs, of which 210 are tumor cases.
The tumor cases contain metastases of breast cancer in lymph nodes, ranging from large metastatic areas to small to individual cancerous cells in lymph node tissue.
All non-tumor WSIs are control cases exhibiting no pathological findings.

Preparatory to patch extraction, we subtracted the background as described in \cite{litjens16} to ensure that only patches from foreground regions are sampled. To speed up this process for the large CAMELYON dataset, background subtraction is done at a level where each extracted patch is represented by a single pixel with a value obtained through interpolation.
Then we covered the entire WSI with a regular grid of patch center points and extracted a patch along with its label according to the grid center point.
Thereby, we ensure that the extracted patches exhibit a random portion of overlap of classes, which should help to reduce over-fitting on large homogeneous regions.
After downscaling all extracted patches to match the physical resolution of the BACH data set, we obtained in total 274,272 image patches of physical size $\SI{132}{\micro\metre} \times \SI{132}{\micro\metre}$ at $157 \times 157$ pixels from both CAMELYON data sets.
For pre-training the network, we randomly split the data with a ratio of 80\% and 20\% for training and validation, respectively, making sure that data from all sub-groups of the two challenges is equally represented in training and validation.
This way, we obtained 119,705 normal and 101,347 invasive patches for training and 30,240 normal and 22,980 invasive patches for validation.

For pre-training the network, we used a uniform Xavier initialization and trained the network for 90 epochs starting with a learning rate of \num{1e-3}, which is decreased by a factor of $0.5$ every 20 epochs.
The employed data augmentation strategy is identical to the one used for fine tuning, which is described in Sec.~\ref{sec:partB}.
\subsection{Classification of Histological Images (A)}
\label{sec:partA}
\subsubsection{Data Preparation, Scale Selection and Augmentation}
The data of part A consists of 400 images of size $2048 \times 1536$ pixels with a pixel resolution of $\SI{0.42}{\micro\metre} \times \SI{0.42}{\micro\metre}$.
The images have been assigned one of the four aforementioned classes, if two medical experts agreed to the predominant type of cancer in each image.
For classification, we rescaled all images by a factor of 10 resulting in images of size $205 \times 154$ pixels.

It is important to note that this data set contains multiple subsets of images acquired from the same patient.
In order to evaluate a classification method in a clinically correct way, it is essential to prevent images from the same patient being present in both training and validation set.
Due to the limited amount of data, however, we decided to explicitly drop this constraint and performed a 5-fold cross-validation with a data distribution of 80\% and 20\% for training and validation, respectively.
This way, we wanted to ensure that the network has seen the maximum variability of the data during training.

As pathological images do not have a canonical orientation, we used arbitrary rotations as well as horizontal and vertical flips for data augmentation.
In order to achieve slight scale-invariance we also used random scale changes in the range $[0.5, 2.0]$.
To achieve robustness against spatial recognition of features, we employed random shifting of up to 50\% of the width and height for each image.
Pixels outside of the range of the original image are replaced by their nearest neighbors inside the image.
Finally, we normalized each image by the mean and the standard deviation of all images in the data set. 
\subsubsection{Network Architecture and Training}\label{sec:densenet}
We used a DenseNet-161 architecture as proposed by Huang \etal \cite{huang17} which generalizes the concept of residual learning introduced by He \etal \cite{he16}.
The architecture consists of seven stages, six spatial reduction stages and one classifier stage.
Each spatial reduction uses a stride of $2\times2$.
The first two stages consist of a single convolution (kernel size $7\times7$) and a single max pooling (kernel size $3\times3$), respectively.
The next four stages consist of densely connected convolutions (kernel sizes $3\times3$ and $1\times1$) followed by a full $2\times2$ max pooling.
The head of the classifier consists of a global average pooling of the spatial feature map and a single fully-connected layer.
We employ a categorical cross-entropy loss and retain the weights with highest classification accuracy on the validation set.
Neither dropout nor weight-decay were used.

For transfer learning, we first trained only the fully connected layer for 25 epochs with a learning rate of \num{1e-3} in order to avoid over-fitting.
Next, we trained the whole network for 250 epochs with a learning rate of \num{2e-4}.
One epoch consists of all possible batches of size 32 of the training data.
The training data was randomly shuffled between each epoch.
\subsection{Segmentation of Whole Slide Images (B)}
\label{sec:partB}
\subsubsection{Data Preparation, Scale Selection and Augmentation}
The data of part B consists of 30 whole slide images of which only 10 are annotated with regard to the aforementioned tissue classes.
All WSIs have a spatial resolution of $\SI{0.467}{\micro\metre} \times \SI{0.467}{\micro\metre}$ per pixel.

From the ten annotated WSIs, we extracted patches of physical size $\SI{330}{\micro\metre} \times \SI{330}{\micro\metre}$ into $157 \times 157$ pixels, corresponding to a down-sampling factor of $4.5$.
We followed a similar procedure as described in Sec.~\ref{sec:camelyon}. The only difference is that the background subtraction is done at the patch-level, with a patch being considered background if at least 80\% of its pixels are considered background.
We obtained a total of 24,406 patches, with 13,280 being labeled normal, 903 benign, 354 in situ, and 9,869 invasive.
Due to the very limited amount of data in the benign and the carcinoma in situ classes, we refrained from splitting training and validation data according to the individual WSIs and performed a random splitting, as done for part A of the challenge, and refrained from performing a cross-validation.
Again, our motivation was to expose the maximum variability to the network during training.

We employed a similar strategy for data augmentation as described in Sec.~\ref{sec:partA}:
The main difference is that missing pixels are replaced by the actual pixels from the larger image, except on the borders of the WSI where they are replaced by their nearest neighbors.
Finally, in order to achieve robustness with respect to color perturbations introduced by varying staining conditions for example, we employed the color augmentation procedure suggested by \cite{liu17}.
All other augmentation parameters are kept as in Sec.~\ref{sec:partA}

\subsubsection{Additional Data}
To further reduce data shortage, we added additional data by partially annotating 16 of the 20 originally non-annotated WSIs.
This was done with the help of a trained pathologist.
In particular, we aimed at reducing the problem of imbalanced classes and specifically annotated regions containing benign malformations and carcinoma in situ.

After performing the data extraction again as described above, we extracted a total of 41,506 image patches, with 25,230 normal, 1,723 benign, 1,759 in situ and 12,794 invasive tissue regions.


%
\subsubsection{Network Architecture and Training}
We used the same DenseNet-161 architecture as in part A, as described in Sec.~\ref{sec:densenet}.
For transfer learning, we first trained only the fully connected layers for 6 epochs with a learning rate of \num{5e-3} in order to avoid over-fitting.
Next, we trained the whole network for 60 epochs with a learning rate of \num{1e-3} and 40 epochs with a learning rate of \num{5e-4}.
In order to compensate for the highly imbalanced classes, we employed log-balanced class weights, \ie the weight for class $c$ is defined as $\log(N / N_c)$, where $N$ denotes the number of all training patches and $N_c$ the number of training patches belonging to class $c$.
\subsubsection{Patch-based Segmentation and Post-Processing}
In order to produce a segmentation of a full WSI, we first down-scale the WSI to obtain the expected resolution of the classifier.
We then classify every grid center point of a grid with cell size $32 \times 32$ pixels.
In total, the down-sampling factor is approximately $144$.

The resulting label image is then post-processed by applying a median filter to smooth the segmentation and a small dilation of all non-normal classes (overlapping classes are resolved in the order benign $<$ in-situ $<$ invasive) to slightly decrease the false-negative rate and slightly increase the size of tumor regions after decreasing them with the median filter.

\section{Evaluation}
\label{sec:evaluation}
We conducted several experiments for both parts of the challenge on a dedicated workstation with Intel i7-6850K processor, 64GB RAM and two NVIDIA Geforce GTX 1080 Ti graphics cards.
As an operating system we used Ubuntu 16.04 LTS, endowed with docker and NVIDIA-docker.
For implementing the network architecture and conducting the training we used python 2.7.12, keras 2.1.3 and tensorflow 1.4.0 backend (official tensorflow docker).
Training time on this machine (using one GPU) was around 10 hours for pre-training on the CAMELYON data set and between 7 and 9 hours for transfer learning.
The inference time per image or patch is around \SI{37}{\milli\second} and for a full WSI is around \SI{30}{\minute}.
\subsection{Classification of Histological Images (A)}
The results of our experiments for part A of the BACH challenge are summarized in Tab.~\ref{tab:partA}.
In lines one and two, we report the achieved accuracies for baseline approaches based on the VGG-19 and Inception-v3 architectures \cite{simonyan2014,szegedy16}
The DenseNet-161 architectures, which were trained using the same hyper-parameter settings as described in Sec.~\ref{sec:partA}, but less aggressive data augmentation.
The performed experiments show that the baseline architectures exhibit worse performance in our setting.
Although we did not perform a grid search for hyper-parameter tuning, we believe that the discrepancy between these architectures is not solely caused by a discrepancy in quality of the hyper-parameters, such that the reported results give a fair qualitative impression of the performance of these architectures.
In line three and four of Tab.~\ref{tab:partA}, we report the results of the proposed approach using ImageNet-data and CAMELYON data for pre-training, respectively.
These two experiments suggest that pre-training using ImageNet-data seems to outperform pre-training on CAMELYON data.
For composing the challenge submission, we selected the best performing network pre-trained on ImageNet from the cross-validation experiment, \cf line three of Tab.~\ref{tab:partA}.
\begin{table}[t]
	\centering
	\begin{tabular}{cccc}
		\toprule
		\textbf{architecture} & \textbf{pre-training on} & \textbf{splitting} & \textbf{accuracy} \\ \midrule
        \textbf{VGG-19} & ImageNet & 5-fold  cross-validation & 92.5\% \\ \midrule
        \textbf{Inception-v3} & ImageNet & 5-fold  cross-validation & 91.25\% \\ \midrule \midrule
        \textbf{DenseNet-161} & ImageNet & 5-fold  cross-validation & \textbf{94\%} \\ \midrule
		\textbf{DenseNet-161} & CAMELYON & 5-fold cross-validation & 76.75\% \\
		 \bottomrule
	\end{tabular}
	\caption{\textbf{Classification Accuracy of Various Trainings (part A):} Baseline approaches (in the first two rows) are compared with less aggressive data augmentation but same hyper-parameters (DA-) and the DenseNet-161 architecture tested with different pre-trainings (last two rows).}
	\label{tab:partA}
\end{table}
\subsection{Segmentation of Whole Slide Images (B)}
For the second part of the challenge we conducted two main experiments:
At first, we limited ourselves to the 10 annotated WSIs, using a random stratified split into 80\% training and 20\% validation data.
We tested this approach using networks pre-trained on CAMELYON as well as pre-trained on ImageNet.
Secondly, we added additional data from selected WSIs as described in Section~\ref{sec:partB} and repeated the training, comparing the obtained results with VGG-19 and Inception-v3 architectures as a baseline approaches, \cf Tab.~\ref{tab:partB}.

Similar to part A, the DenseNet architecture outperforms the Inception architecture and pre-training on ImageNet outperforms pre-training on CAMELYON.
We chose the model trained on the extended data set for submission as it achieves highest accuracy.
Since the remaining four unlabeled WSIs do not exhibit sufficient variability in order to assess the network performance based on the score suggested by the challenge, we based our decision solely on patch-accuracy.

\begin{table}[t]
	\centering
	\begin{tabular}{cccc}
		\toprule
		\textbf{architecture} & \textbf{pre-training on} & \textbf{data} & \textbf{patch-based acc.} \\ \midrule
		\textbf{DenseNet-161} & ImageNet & annotated 80/20 split & 95.75\% \\ \midrule 
		\textbf{DenseNet-161} & CAMELYON & annotated 80/20 split & 95.33\% \\ \midrule \midrule
		\textbf{VGG-19} & ImageNet & ext. annotated 80/20 split & 96.04\%\\ \midrule
		\textbf{Inception-v3} & ImageNet & ext. annotated 80/20 split & 95.51\%\\ \midrule
		\textbf{DenseNet-161} & ImageNet & ext. annotated 80/20 split & \textbf{96.24\%}\\
		\bottomrule
	\end{tabular}
	\caption{\textbf{Patch-Based Classification Accuracy of Various Trainings (part B)}: Comparison of different architectures, pre-trainings and datasets w.r.t. patch-accuracy on WSIs.}
	\label{tab:partB}
\end{table}

%
%


%
\section{Discussion}
\label{sec:discussion}
%
Regarding the results for part A of the challenge, it becomes apparent that the DenseNet architecture outperforms the other baseline methods.
More interesting than this first qualitative comparison, however, is the fact that pre-training on ImageNet is considerably better than pre-training on CAMELYON data.
We hypothesize that this discrepancy arises from the fact that features learned from the CAMELYON data base do not generalize well enough to the specific appearance of the images from part A.

Comparing the achieved results to the ones reported by Araujo \etal \cite{araujo17} is not straightforward: In \cite{araujo17} a classification accuracy of 78\% for the same task is reported, however, the used dataset is even smaller (285 images) and we do not have any information regarding the splitting of the data.


Regarding the results for part B, we observe again that the DenseNet architecture outperforms the baseline approaches, \ie the VGG-19 and Inception-v3 architectures.
Furthermore, we observed comparable results for the DenseNet trained on the original ten WSIs and the extended data base of 26 WSIs.
As we observed a better generalization performance in preliminary experiments, where we gradually added additional training data, we decided to submit the network which has seen the largest data variability during training to the challenge phase.

In both parts, we observed a better performance for networks pre-trained on ImageNet data in comparison to the ones pre-trained on CAMELYON data.
By training networks on the CAMELYON data base we were hoping to learn features, particularly in the first layer, which are better suited to digital pathology images.
On the other hand, networks pre-trained on ImageNet are known to learn very robust and general features due to the high variability of the ImageNet data base and it might be that the features learnt from the CAMELYON data base generalize less well to the data of this challenge.

\section{Conclusion}
\label{sec:conclusion}
The conducted experiments demonstrate that densely connected convolutional networks are well-suited for transfer learning, even in case the considered data set is small.
In order to develop classification algorithms which can be used in clinical practice, a significantly larger amount of data is necessary.
We want to emphasize that the chosen splittings for training and validation (in both part A and B) are not suited for a clinical evaluation.
In addition to this, a data base for training such a network possibly requires more precise and also different annotations.
This cannot only be observed by inspecting the rather coarse annotations of the WSIs, but also by the fact that part A only contains images where two pathologists agreed.
In fact, computer-assisted diagnoses would be particularly helpful in those excluded cases.
However, future work should not be limited to the creation of larger and carefully annotated data bases.
The development of sophisticated feature visualization techniques will be crucial to not only understand performance differences of differently trained networks, but also to make the computed decision more understandable to the medical expert.  
\bibliographystyle{splncs03}
\bibliography{references}

\end{document}